\documentclass[letterpaper,10pt,conference]{ieeeconf}

\IEEEoverridecommandlockouts
\usepackage{amsmath,amssymb}
\usepackage{graphicx}
\usepackage{capt-of}
\usepackage{booktabs}
\usepackage{multirow}
\usepackage{xspace}
\usepackage{microtype}
\usepackage{hyperref}

\newcommand{\method}{\textsc{HIL-UMI}\xspace}

\title{\LARGE \bf
HIL-UMI: Bringing Human-in-the-Loop Post-Training of Vision-Language-Action Models to Universal Manipulation Interface
}

\author{Zimu Han$^{*1,4}$, Yiming Zeng$^{*1,4}$, Jiyao Zhang$^{*\ddagger1,2,3}$, Zihao Zhao$^{1}$, Yuanfei Wang$^{1,2,3}$, Yixiang Jin$^{5}$ \\ Shiqi Li$^{5}$, Shuangben Chen$^{1}$, Wei Huang$^{1}$, Ruodai Li$^{5}$, Hui Shen$^{5}$ and Hao Dong$^{\dagger1,2,3}$
\thanks{* Equal contribution. $\ddagger$ Project lead. }
\thanks{$\dagger$ Corresponding author. Correspondence to \texttt{hao.dong@pku.edu.cn}. }
\thanks{$^{1}$ Center on Frontier Computing Studies, School of Computer Science, Peking University, China, $^{2}$National Key Laboratory for Multimedia Information Processing, School of Computer Science, Peking University, China, $^{3}$PrimeBot, China, $^{4}$Xi'an Jiaotong University, China, $^{5}$JD Technology, China}
}

\IEEEaftertitletext{%
    \begin{minipage}{\textwidth}
        \centering
        \includegraphics[width=\textwidth]{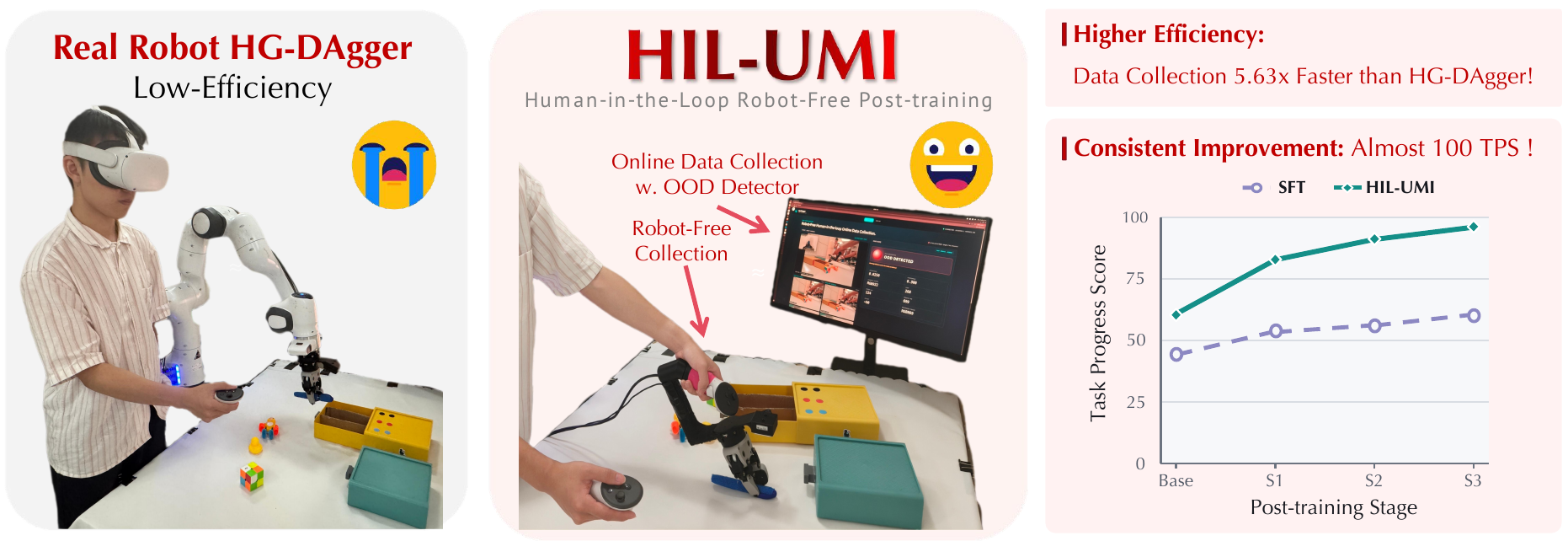}
        \captionof{figure}{\textbf{Human-in-the-Loop Robot-Free Post Training.} Real-robot HG-DAgger requires policy rollouts and human intervention on the robot, resulting in low collection efficiency. In contrast, HIL-UMI performs policy-guided, robot-free data collection with an online OOD detector, achieving higher performance with 5.63× faster data collection. }
        \label{fig:teaser}
        \vspace{0.25em}
    \end{minipage}%
}

\begin{document}

\maketitle
\thispagestyle{empty}
\pagestyle{empty}

\begin{abstract}
Large-scale vision-language-action (VLA) models provide powerful priors for robot manipulation, yet adapting them to a specific deployment remains challenging. Supervised fine-tuning (SFT) on task-specific demonstrations provides a step toward deployment, but faces two persistent limitations: static data provide limited coverage of out-of-distribution states, and standard imitation objectives do not distinguish progressing behavior from less useful data. Interactive post-training can address these limitations, but typically requires repeated policy execution and human intervention on a physical robot. We introduce HIL-UMI, a policy-guided Universal Manipulation Interface (UMI) framework for robot-free human-in-the-loop VLA post-training. During handheld UMI demonstrations, HIL-UMI queries the current policy on the same observation stream without executing its predictions. The Energy Score compares the human action trajectory with policy inference and triggers collection when their discrepancy indicates an out-of-distribution region. In a separate feedback loop, low online advantage predictions identify essential segments for refining a progress-based advantage estimator. The updated estimator then guides advantage-conditioned behavioral cloning using a balanced mixture of base demonstrations and new policy data. This design preserves the iterative and policy-aware nature of human-in-the-loop learning while decoupling data collection from robot deployment. Experiments on four real-world tasks spanning long-horizon and precise manipulation show that HIL-UMI achieves consistent improvement over SFT and benefits from both targeted collection and advantage refinement. Moreover, HIL-UMI outperforms HG-DAgger on Clean Up Table with lower per-frame collection time, suggesting a scalable path for VLA post-training across operators and locations. Project page: \url{https://hil-umi.github.io}. 
\end{abstract}

\section{Introduction}
\label{sec:introduction}

Large-scale vision-language-action (VLA) models~\cite{kim2025openvla,black2025pi0,black2025pi05,zhang2026joyaira} acquire broad manipulation priors from diverse robot and vision--language data~\cite{ghosh2024octo,contributors2025agibotdigitalworld,zheng2026worldline,wu2026zimablue}, providing strong initializations for downstream robot learning. Beyond acquiring individual behaviors, such pretraining enables skill reuse across tasks and generalization across objects and scenes. Yet broad competence does not guarantee reliable execution in a particular deployment: the target embodiment, observation setup, workspace, dynamics, and required precision can differ from those seen during pretraining. Task-specific post-training is therefore a critical bridge between general-purpose representations and robust closed-loop behavior in real-world deployments~\cite{kim2025oft,chen2025conrft,zang2026rluxvla,pan2026sop,yu2026correctnav,zhou2026evowam}. It adapts a pretrained policy to concrete operating conditions and is especially important for long-horizon and precise manipulation, where small local errors can determine overall task success~\cite{yu2026chi0,zhang2026hipolicy,zhao2023act,long2025checkmanual,long2026scalingmanual,wu2025canonical3dtactile,li2025adaptivevisuotactile}.

The dominant paradigm for this adaptation collects task demonstrations on physical robots and applies supervised fine-tuning (SFT)~\cite{kim2025oft,mandlekar2022whatmatters}. However, SFT leaves two core problems unresolved. First, behavioral cloning is susceptible to covariate shift and compounding errors: small errors lead the policy to states poorly covered by static demonstrations~\cite{ross2011dagger,laskey2017dart}. Static demonstrations mainly cover expert-visited states, so more nominal data may still miss the out-of-distribution (OOD) states reached by the learned policy. 
Second, SFT weights all demonstration samples equally, regardless of their contribution to task progress~\cite{peng2019awr,nair2021awac,wang2020crr,kostrikov2022iql}. 

Human-in-the-loop post-training addresses these limitations more directly~\cite{spencer2020interventions,cai2025aim,chen2025conrft}. DAgger queries expert actions at learner-visited states and aggregates them into the training set, directly expanding coverage to states induced by policy errors~\cite{ross2011dagger}. Related human-gated~\cite{kelly2019hgdagger,mandlekar2020iwr,hoque2022thriftydagger,liu2025sirius} and real-robot variants~\cite{luo2025hilserl,hoque2023fleetdagger,jiang2025transic} similarly focus on teleoperation, rollouts, interventions, or reward feedback around policy failures. Building on this interactive paradigm, RL with Experience and Corrections via Advantage-conditioned Policies (RECAP) incorporates demonstrations, autonomous on-robot experience, and expert teleoperated corrections into advantage-conditioned policy training~\cite{physicalintelligence2025pistar06,peng2019awr,yu2026chi0,yang2026aloe}. On-policy experience exposes learner-induced OOD states, while advantage conditioning distinguishes data utility; RECAP thereby addresses both problems and achieves strong performance on challenging real-world tasks. This success, however, requires repeated physical-robot deployment, making collection expensive and difficult to parallelize across operators and locations~\cite{hoque2023fleetdagger,pan2026sop}. Teleoperation also makes long-horizon and high-precision demonstrations difficult to collect at scale~\cite{dass2023pato}.

To remove this dependency, we draw inspiration from the Universal Manipulation Interface (UMI), whose portable, low-cost handheld grippers collect robot-compatible observations and actions without access to the target robot~\cite{chi2024umi,zhaxizhuoma2025fastumi,ha2025umionlegs}. Direct hand demonstrations let operators express complex and precise behaviors naturally, while portability enables collection across operators and locations~\cite{wei2026hifiumi}. 

We therefore propose \method, a UMI-based human-in-the-loop framework for robot-free VLA post-training, where a human demonstrates the task while the current policy predicts actions from the same observation stream without executing those predictions on a robot. The discrepancy between policy inference and human demonstration determines whether the current state is OOD and whether additional data should be collected here. This design combines policy-conditioned feedback with robot-free, parallelizable collection, extending UMI collection to target the current policy's blind spots and improve its training data.

Specifically, in each round, we collect two separate UMI datasets for distinct purposes. For policy post-training, we repeatedly run the current policy on the UMI observation stream and retain segments whose human actions deviate from the policy's trajectory distribution beyond a threshold as OOD data. Separately, we run the advantage model online and collect dedicated training data whenever its predicted advantage falls, treating these low-scoring segments as hard cases for improving the advantage model. We update the advantage model with this second dataset, then perform advantage-conditioned behavioral cloning (ACBC) on a mixture of the base and collected OOD data.

Our contributions are threefold:
\begin{itemize}
    \item We introduce a UMI-based human-in-the-loop framework that moves iterative VLA post-training off the robot, decoupling policy improvement from physical deployment and opening a path toward scalable, parallel data collection across operators and locations.
    \item We propose a real-time OOD detection method that compares human action trajectories against the policy's trajectory distribution during UMI collection, together with an iterative workflow that updates an advantage model and performs ACBC in every round.
    \item We validate the framework on four challenging long-horizon or precise tasks. Results show that HIL-UMI achieves substantial gains over SFT, better performance and higher collection efficiency than  HG-DAgger.
\end{itemize}

\begin{figure*}[t]
    \centering
    \includegraphics[width=\textwidth]{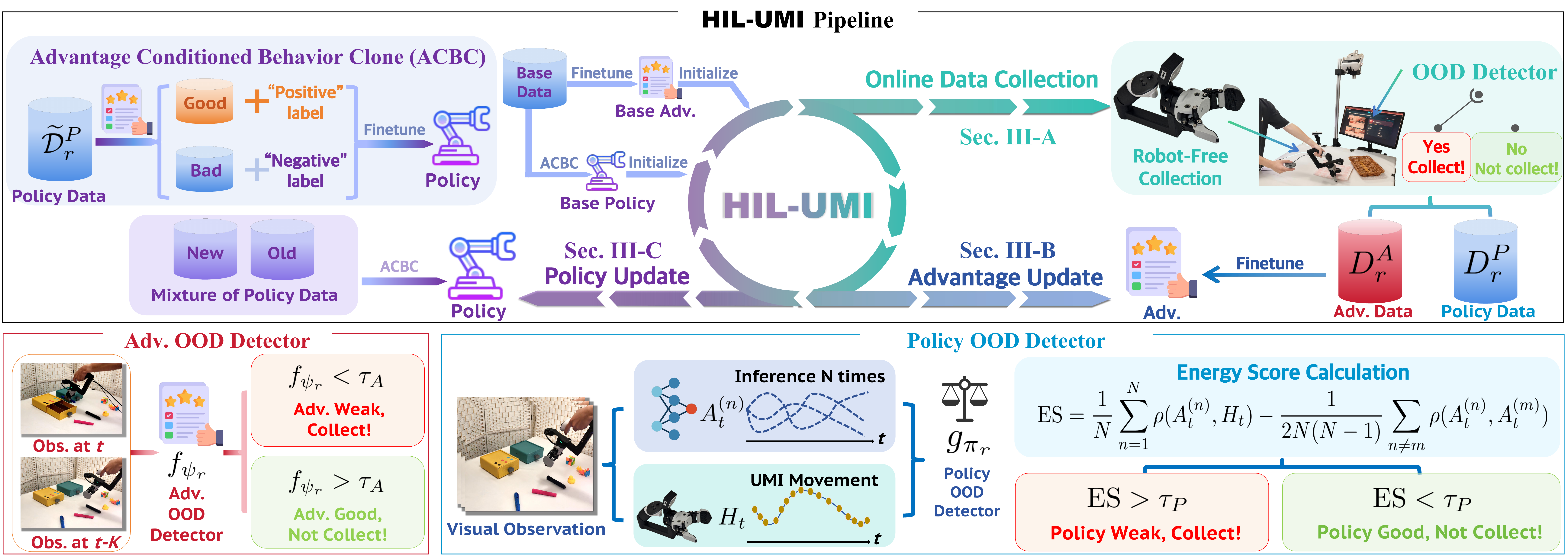}
    \caption{\textbf{Overview of HIL-UMI.} In each round, we first run advantage OOD detector online to collect advantage data. We then run the current policy OOD detector and collect policy data from segments. Next, we update the advantage estimator then using it to construct the advantage-labeled training dataset and update policy through ACBC.}
    \label{fig:method_overview}
\end{figure*}
\section{Related Work}
\label{sec:related}

\subsection{VLA Post-Training and Interactive Policy Improvement}
Supervised fine-tuning (SFT) adapts pretrained VLA policies, yet expert data offer limited state coverage, leaving policies vulnerable to compounding errors~\cite{ross2011dagger,laskey2017dart,kim2025oft,mandlekar2022whatmatters}. Existing methods differ in the source of corrective signals~\cite{chen2025conrft,zang2026rluxvla}. On the static-data side, GR-RL uses offline reinforcement learning to estimate task progress and filter suboptimal demonstrations~\cite{li2025grrl}, while $\chi_0$ combines model arithmetic and stage advantage to reconcile heterogeneous data distributions~\cite{yu2026chi0}. Interactive deployment methods instead obtain supervision from learner-induced experience: DAgger queries expert actions at states visited by the learner and aggregates them into the training set~\cite{ross2011dagger,spencer2020interventions,hoque2023fleetdagger,cai2025aim}; HIL-SERL couples real-robot reinforcement learning with real-time human interventions~\cite{luo2025hilserl,jiang2025transic,chen2025conrft}; and RECAP trains $\pi_{0.6}^{*}$ from demonstrations, autonomous on-robot experience, and teleoperated corrections~\cite{physicalintelligence2025pistar06,pan2026sop}. Despite their effectiveness, their final correction or alignment still relies on physical-robot rollouts or interventions, incurring hardware and operator costs and limiting parallel scaling across users and locations~\cite{hoque2023fleetdagger,dass2023pato,pan2026sop}.

\subsection{Robot-Free Data Collection with UMI}
UMI replaces robot teleoperation with a portable handheld gripper that records robot-compatible observations and actions, enabling in-the-wild teaching and deployment across robot embodiments~\cite{chi2024umi,ha2025umionlegs}. FastUMI simplifies the hardware and deployment stack to support scalable, robot-independent collection~\cite{zhaxizhuoma2025fastumi}, while MV-UMI adds a third-person view to provide richer spatial context and mitigate cross-embodiment observation shift~\cite{rayyan2025mvumi}. HiFi-UMI further improves trajectory fidelity, bimanual relative-pose estimation, synchronization, and field of view, showing that UMI-only post-training can approach the performance of real-robot teleoperation~\cite{wei2026hifiumi}. However, its real-time feedback targets sensing and capture quality rather than predictions from the current policy. Consequently, collection remains centered on data fidelity and general coverage rather than policy-conditioned selection of demonstrations that address the policy's specific blind spots.

\subsection{UMI-Based Human-in-the-Loop Post-Training}
Recent work has begun to close the loop between UMI collection and policy improvement. RoboPocket visualizes predicted policy trajectories to solicit robot-free corrections and fine-tune the policy, but weakness identification relies on human interpretation and its learning objective does not estimate the utility of individual segments~\cite{fang2026robopocket}. EgoGuide uses dataset-level visual-geometric novelty to guide demonstrations toward under-covered initial states, but its feedback is not conditioned on the current policy and therefore does not directly target policy-specific blind spots~\cite{xu2026egoguide}.  In contrast, our framework detects policy-conditioned OOD segments through trajectory prediction and iteratively learns an advantage model to label selected data by utility during ACBC.

\section{Method}
\label{sec:method}

\label{sec:method_overview}

We consider a task instruction $c$ and an initial UMI~\cite{chi2024umi} dataset
$\mathcal{D}_0=\{\tau_i\}_{i=1}^{M}$, where
$\tau_i=\{(o_{i,t},a_{i,t})\}_{t=0}^{L_i-1}$ contains synchronized
observations and human actions. 
These base demonstrations are first assigned progress targets and used to initialize a
two-observation advantage estimator $f_{\psi_0}$. We then use $f_{\psi_0}$ to
assign binary advantage labels to the same demonstrations and directly obtain
the base policy by advantage-conditioned behavior cloning
(ACBC),
\begin{equation}
    \theta_0 = \arg\min_{\theta}
    \mathbb{E}_{(o,a,b)\sim\mathcal{D}_0}
    \left[\ell_{\mathrm{BC}}
    \bigl(\pi_{\theta}(\cdot\mid o,c^b),a\bigr)\right],
    \label{eq:base_policy}
\end{equation}
where $\ell_{\mathrm{BC}}$ denotes the native action-prediction loss of the
policy and $b$ is the advantage label. The progress supervision and ACBC are detailed in
Secs.~\ref{sec:advantage_refinement}
and~\ref{sec:policy_update}.

Post-training proceeds for rounds $r=1,\ldots,R$ with separate datasets for
advantage refinement and policy improvement. Let $\mathcal{D}^{A}_r$ and
$\mathcal{D}^{P}_r$ denote the data newly collected for these two purposes in
round $r$, respectively. In each post-training round, we allocate equal frame budgets to the two collection streams, such that \(|D_r^A|=|D_r^P|\).  We define $\mathcal{D}^{P}_0=\mathcal{D}_0$ and let
$\mathcal{D}^{A}_0$ be its progress-labeled version. 
In each round, we first run \(f_{\psi_{r-1}}\) online to collect \(\mathcal{D}^{A}_{r}\). We then run the current policy \(\pi_{\theta_{r-1}}\) alongside a human UMI demonstration and collect \(\mathcal{D}^{P}_{r}\) from segments where the policy’s action distribution is identified as OOD. Next, we update the advantage estimator by continuing training from \(\psi_{r-1}\) on a balanced training mixture, obtaining \(\psi_r\). Finally, using \(f_{\psi_r}\), we construct the advantage-labeled training dataset and continue ACBC from \(\theta_{r-1}\) to obtain \(\theta_r\). Figure~\ref{fig:method_overview} summarizes this data-collection and model-update loop.

\subsection{Online Data Collection}
\label{sec:online_ood}


\noindent{\textbf{OOD Definition.}} In this work, we use OOD to describe task situations poorly represented in the training data of the corresponding model. We flag OOD through inconsistencies between model predictions and human demonstrations: a large discrepancy between the demonstrated action chunk and the policy’s predicted action distribution, or a low advantage prediction despite demonstrated task progress.

\subsubsection{Policy Data Collection}

At observation $o_t$, the operator produces a human action chunk
$H_t=(h_{t,1},\ldots,h_{t,T})$~\cite{zhao2023act,zhang2026hipolicy}. In
parallel, we perform $N=10$ stochastic policy inferences at the same
observation and instruction,
\begin{equation}
    A_t^{(n)} \sim \pi_{\theta_{r-1}}(\cdot\mid o_t,c),
    \qquad n=1,\ldots,N,
    \label{eq:policy_samples}
\end{equation}
which form an empirical approximation to the policy's action-chunk
distribution~\cite{chi2023diffusionpolicy,black2025pi0}. The human and
predicted chunks are compared only after the corresponding human chunk has
been observed, while policy sampling itself is performed concurrently with
UMI collection.

For pose actions, write the $k$-th action of a chunk as
$a_k=(p_k,R_k,g_k)$, comprising end-effector position, orientation, and
gripper command. We compare action chunks $A$ and $B$ in a common coordinate frame
using
\begin{align}
    \rho^2(A,B) = \frac{1}{T}\sum_{k=1}^{T} \bigl[
        &\lambda_p\lVert p_k^A-p_k^B\rVert_2^2
        + \lambda_R d_R(R_k^A,R_k^B)^2 \nonumber\\
        &+ \lambda_g\lVert g_k^A-g_k^B\rVert_2^2
    \bigr],
    \label{eq:chunk_distance}
\end{align}
where the coefficients $\lambda_p$, $\lambda_R$, and $\lambda_g$ balance the
action components. To measure the difference between orientations, we use the
geodesic distance on $\mathrm{SO}(3)$,
\begin{equation}
    d_R(R_1,R_2) = \arccos
        \!\left(
        \frac{\operatorname{tr}(R_1^\top R_2)-1}{2}
        \right).
    \label{eq:rotation_distance}
\end{equation}

We represent the discrepancy between the single human chunk and the predicted
distribution with the empirical Energy Score~\cite{gneiting2007strictly,gneiting2008multivariate,szekely2013energy}:
\begin{align}
    \operatorname{ES}(H_t)
    ={}& \frac{1}{N}\sum_{n=1}^{N}\rho(A_t^{(n)},H_t) \nonumber\\
       &-\frac{1}{2N(N-1)}
       \sum_{n\neq m}\rho(A_t^{(n)},A_t^{(m)}).
    \label{eq:energy_score}
\end{align}
The first term measures how far the policy samples lie from the human action. The second accounts for the dispersion of the policy samples and prevents the criterion from reducing to an average pointwise error. Their balance makes the score sensitive to both location and predictive spread: excessive spread raises the sample-to-human distances, whereas a collapsed distribution away from the human action receives no diversity correction. The score requires neither a Gaussian assumption nor an explicit likelihood, making it suitable for flow-based policies. A large Energy Score indicates strong disagreement between the demonstrated action and the policy’s predictive distribution; we operationally treat the corresponding state region as OOD.

We consequently define the HIL-UMI OOD detector as
\begin{equation}
    \delta_t^{P}=\mathbb{I}\!\left[
        \operatorname{ES}(H_t)>\tau_{P}
    \right],
    \label{eq:ood_trigger}
\end{equation}
where $\tau_{P}$ is the policy OOD detection threshold shared across all tasks. When a completed chunk triggers $\delta_t^{P}=1$, the operator records an expert demonstration from the current state until the current subtask is completed. Repeating this procedure yields
$\mathcal{D}^{P}_r$, which concentrates the policy update on states where the current action distribution does not cover the human solution without requiring policy execution.

\subsubsection{Advantage Data Collection}

During advantage-data collection in round $r$, 
we estimate relative task progress from an observation pair with a $K$-frame temporal offset 
whenever $t\geq K$, as follows:
\begin{equation}
    \widehat{A}^{\mathrm{online}}_t
    =f_{\psi_{r-1}}(o_{t-K},o_t,c),
    \qquad
    \delta_t^{A}=\mathbb{I}\!\left[
        \widehat{A}^{\mathrm{online}}_t<\tau_{A}
    \right],
    \label{eq:advantage_trigger}
\end{equation}
where \(\tau_A\) is calibrated per task using the initial advantage estimator \(f_{\psi_0}\). We evaluate \(f_{\psi_0}\) on observation pairs separated by \(K\) frames from the task’s base dataset \(D_0\). Let \(\kappa_0\) denote the empirical cutoff selecting the top \(\eta\) fraction of predicted advantages (\(\eta=0.3\)), which is also used for base-data advantage labeling in Sec.~\ref{sec:policy_update}. We set \(\tau_A=\kappa_0/2\), adapting the collection threshold to differences in progress scale across tasks over the fixed temporal interval. Assuming the operator is demonstrating task-progressing behavior, predictions below this calibrated threshold identify segments where the estimator may underestimate task progress.
Upon such a trigger, the operator records a new demonstration from the current state through the end of the subtask.
These segments form $\mathcal{D}^{A}_r$ and receive the progress
labels described in Sec.~\ref{sec:advantage_refinement}.

\begin{figure*}[!t]
    \centering
    \includegraphics[width=\textwidth]{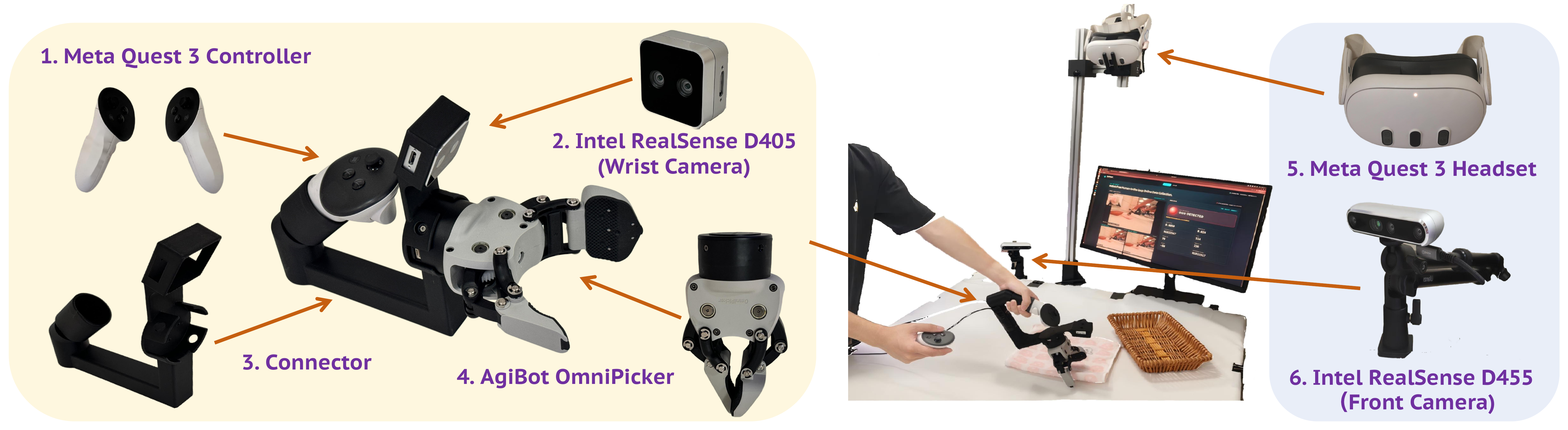}
    \caption{\textbf{Hardware setup.} Components of the custom UMI device and
    the corresponding data-collection setup.}
    \label{fig:hardware_setup}
\end{figure*}

\subsection{Advantage Model Training}
\label{sec:advantage_refinement}

Rather than deriving relative task progress from the difference of two independently predicted values, we train $f_\psi(o_u,o_v,c)$ to predict relative task progress from $o_u$ to $o_v$, which can reduce the compounding error of estimation~\cite{yu2026chi0}. To accommodate both full base episodes and the segmental episodes collected in post-training rounds, we design a linear progress target. For base episodes, the progress target is:  
\begin{equation}
    z_{i,t}=\frac{t}{L_i-1},
    \qquad t=0,\ldots,L_i-1,
    \label{eq:base_progress}
\end{equation}
and we define the average base-episode length as
\begin{equation}
    \overline{L}_0=\frac{1}{M}\sum_{i=1}^{M}L_i.
    \label{eq:mean_base_length}
\end{equation}
An iterative segment $\sigma_{r,j}$ has length $L_{r,j}$ and is intentionally
terminated when its current subtask is completed. Since it is not a complete
episode, assigning it the full range $[0,1]$ would overstate its progress.
Instead, we use
\begin{equation}
    z_{r,j,t}=\frac{t}{L_{r,j}-1}
              \frac{L_{r,j}}{\overline{L}_0},
    \qquad t=0,\ldots,L_{r,j}-1,
    \label{eq:partial_progress}
\end{equation}
so that the segment spans $0$ to $L_{r,j}/\overline{L}_0$ and has a temporal
progress scale consistent with the base data.

For two distinct frames $u$ and $v$ uniformly sampled from the same trajectory
or segment, the signed regression target is $y_{u,v}=z_v-z_u$. Sampling
multiple temporal spans during training, we optimize the following objective:
\begin{equation}
    \mathcal{L}_{A}(\psi)
    =\mathbb{E}_{(o_u,o_v,y_{u,v})}
    \left[\bigl(f_\psi(o_u,o_v,c)-y_{u,v}\bigr)^2\right].
    \label{eq:advantage_loss}
\end{equation}

To balance hard cases found in the current round against data collected
before, we define the mix operator
\begin{equation}
    \operatorname{Mix}(\mathcal{D}_{\mathrm{new}},\mathcal{D}_{\mathrm{hist}})
    \triangleq \alpha \cdot \operatorname{Unif}(\mathcal{D}_{\mathrm{new}})
    + (1-\alpha) \cdot \operatorname{Unif}(\mathcal{D}_{\mathrm{hist}}),
    \label{eq:balanced_replay}
\end{equation}
where $\operatorname{Unif}(\mathcal{D})$ is uniform sampling from a dataset,
and $\alpha$ is the coefficient balancing current-round and historical data.
The dataset for updating advantage is
\begin{equation}
    \widetilde{\mathcal{D}}^{A}_r
    = \operatorname{Mix}\!\left(
        \mathcal{D}^{A}_r,
        \bigcup_{j<r}\mathcal{D}^{A}_j
    \right).
    \label{eq:adv_replay}
\end{equation}
Continuing from $\psi_{r-1}$, we minimize Eq.~\eqref{eq:advantage_loss} on
$\widetilde{\mathcal{D}}^{A}_r$ to obtain $\psi_r$.

\subsection{Policy Update with ACBC}
\label{sec:policy_update}

For each post-training round, to balance the historical data and the current-round data, we let the dataset for updating policy be: 
\begin{equation}
    \widetilde{\mathcal{D}}^{P}_r
    = \operatorname{Mix}\!\left(
        \mathcal{D}^{P}_r,
        \bigcup_{j<r}\mathcal{D}^{P}_j
    \right).
    \label{eq:policy_replay}
\end{equation}
The updated estimator scores every eligible sample
in $\mathcal{D}_0$ over the same future horizon,
\begin{equation}
    \widehat{A}_{r,t}=f_{\psi_r}(o_t,o_{t+K},c),
    \qquad t+K<L,
    \label{eq:policy_advantage}
\end{equation}
where $L$ is the containing trajectory length. This definition also applies
to base initialization with $r=0$. Let
$\kappa_r$ be the empirical cutoff that selects the top $\eta\in(0,1)$
fraction of base-data predictions. For each training sample $(o_t,a_t)$, we define
\begin{equation}
    b_{r,t}=
    \begin{cases}
        \mathbb{I}[\widehat{A}_{r,t}\geq\kappa_r],
            & (o_t,a_t)\in\mathcal{D}_0,\\
        1,
            & (o_t,a_t)\in\displaystyle\bigcup_{j=1}^{r}\mathcal{D}^{P}_j.
    \end{cases}
    \label{eq:binary_advantage}
\end{equation}
Thus, base initialization uses only $\mathcal{D}_0$, with $f_{\psi_0}$
assigning positive labels to its top-$\eta$ predictions and negative labels to
the remainder. In later rounds, base samples retain this thresholding rule,
while all newly collected policy samples receive positive labels. We append a
positive label to $c$ when $b_{r,t}=1$ and a negative label otherwise, denoting
the resulting prompt by $c^{b_{r,t}}$. The round-$r$ policy objective is
\begin{equation}
    \mathcal{L}_{\mathrm{ACBC}}(\theta)
    =\mathbb{E}_{(o,a,b)\sim\widetilde{\mathcal{D}}^{P}_r}
    \left[\ell_{\mathrm{BC}}
    \bigl(\pi_\theta(\cdot\mid o,c^b),a\bigr)\right].
    \label{eq:awbc_loss}
\end{equation}
Policy training continues from $\theta_{r-1}$, while inference uses the
positive-advantage prompt to favor task-progressing behaviors. The updated
policy $\theta_r$ and advantage estimator $\psi_r$ are then used in the next
collection round.

\section{Experiments}
\label{sec:experiment}

\subsection{HIL-UMI Hardware Setup}

We collect robot-compatible demonstrations at 30~Hz using the custom UMI device shown in Figure~\ref{fig:hardware_setup}, without executing policy outputs on the physical robot~\cite{chi2024umi}. The device couples an AgiBot OmniPicker gripper to a Meta Quest~3 controller through a custom connector. The Meta Quest~3 headset--controller tracking system measures the device pose in real time, providing the human action trajectories required for online OOD detection. An Intel RealSense D405 mounted on the device captures wrist-view observations, while an Intel RealSense D455 provides a fixed third-person view.

We use a local workstation with NVIDIA RTX 4090D to facilitate real-time policy and advantage inference. In our policy OOD detector implementation, N stochastic policy samples are generated in parallel to decrease inference latency. The latency of policy OOD detector and advantage OOD detector are 112 ms and 93 ms respectively, supporting the human-in-the-loop policy and advantage data collection. 

\subsection{Real-world Experiments}

We evaluate whether HIL-UMI collection improves iterative post-training over conventional demonstration collection, whether advantage refinement provides an additional benefit, and how effectively each method turns human collection time into task progress. Figure~\ref{fig:real_world_tasks} summarizes the four real-world manipulation tasks.

\subsubsection{Real-world Tasks}
\begin{figure*}[!t]
    \centering
    \includegraphics[width=\textwidth]{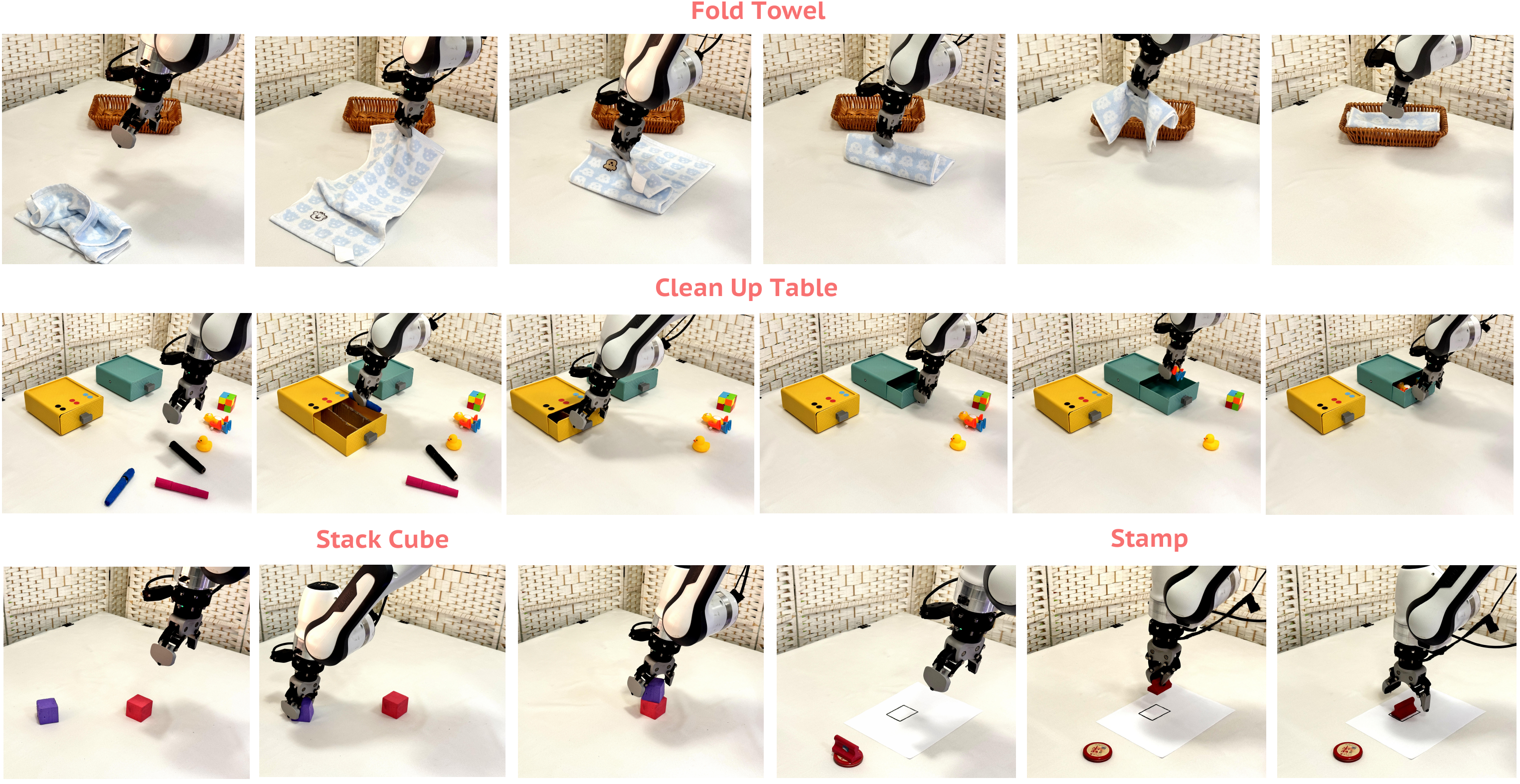}
    \caption{\textbf{Real-world evaluation tasks.} We evaluate our \method on four challenging real-world manipulation tasks with the Franka Panda robot arm.}
    \label{fig:real_world_tasks}
\end{figure*}

We conduct all evaluations on a single Franka arm using the setup shown in
Figure~\ref{fig:real_world_tasks}, and consider the following four tasks:
\begin{itemize}
    \item \textbf{Fold Towel:} This long-horizon task requires the robot to flatten a randomly initialized towel, fold it twice while eliminating wrinkles, and place it in a basket. 
    \item \textbf{Clean Up Table:} The robot first opens the yellow drawer, sorts three pens into
    color-matched slots and closes the drawer. Then it opens the blue drawer, puts away three toys, and closes the drawer. This task features long-horizon manipulation in a housework scenario.
    \item \textbf{Stack Cube:} This precise task requires  the robot to grasp a purple cube and place it on top of a red cube. 
    \item \textbf{Stamp:} The robot grasps a stamp and aligns it inside a marked box on paper. The length and width of the marked box are both $1 \mathrm{cm}$ larger than the stamp body, featuring precise manipulation.
    
\end{itemize}
\subsubsection{Evaluation Protocol}
For a fair comparison, we evaluate each policy checkpoint for 10 trials per task under the same protocol. For each trial, we vary the initial object placement within a $30\,\mathrm{cm}\times60\,\mathrm{cm}$ workspace to evaluate spatial generalization. We report the mean Task Progress Score (TPS), which assigns partial credit to predefined subtasks on a scale from 0 to 100. Table~\ref{tab:tps_scoring_criteria} specifies the complete scoring criteria for each task.

\begin{table}[ht]
    \centering
    \caption{\textbf{Task Progress Score criteria.}}
    \label{tab:tps_scoring_criteria}
    \setlength{\tabcolsep}{2.5pt}
    \renewcommand{\arraystretch}{1.12}
    \footnotesize
    \begin{tabular}{@{}p{0.21\columnwidth}p{0.53\columnwidth}c@{}}
        \toprule
        Task & Subtask & Score \\
        \midrule
        \multirow{4}{*}{Fold Towel}
            & Flatten the towel & $+25$ \\
            & Complete the first fold & $+25$ \\
            & Complete the second fold & $+25$ \\
            & Place the folded towel in the basket & $+25$ \\
        \midrule
        \multirow{4}{*}{Clean Up Table}
            & Place a pen in its color-matched slot & $+10$ each ($\times 3$) \\
            & Put away a toy & $+10$ each ($\times 3$) \\
            & Correctly open the drawer & $+10$ each ($\times 2$)\\
            & Correctly close the drawer & $+10$ each ($\times 2$) \\
        \midrule
        \multirow{3}{*}{Stack Cube}
            & Grasp the purple cube & $+30$ \\
            & Move the purple cube near the red cube & $+30$ \\
            & Place the purple cube on the red cube & $+40$ \\
        \midrule
        \multirow{5}{*}{Stamp}
            & Grasp the stamp & $+20$ \\
            & Move the stamp near the marked box & $+20$ \\
            & Adjust the stamp orientation correctly & $+20$ \\
            & Stamp contact with the paper & $+20$ \\
            & Stamp body inside the marked box & $+20$ \\
        \bottomrule
    \end{tabular}
    \vspace{0.5em}

    \begin{minipage}{0.95\columnwidth}
        \footnotesize
        \raggedright
        \textit{Scoring notes.} For Fold Towel, 5 points are deducted if the towel is wrinkled or misaligned for every subtask.
        For Stack Cube, 15 points are deducted if the robot grasps only one corner of the cube. 
    \end{minipage}
    \vspace{-3pt}
\end{table}

\subsubsection{Training Data and Comparisons}
We adapt the open-source $\pi_{0.5}$ policy~\cite{black2025pi05} to each task. We use a fixed base dataset and a fixed budget for newly collected data in each post-training round. For the long-horizon tasks (Fold Towel and Clean Up Table), the base dataset contains 50 demonstrations, and the per-round data budget is 12,000 frames. For the remaining tasks, the base dataset contains 80 demonstrations, and the per-round data budget is 2,500 frames. 

We compare \method with the SFT baseline based on the data budgets described above. SFT collects conventional UMI demonstrations and applies supervised fine-tuning.  \method refines the advantage estimator and performs advantage-conditioned behavior cloning as described in Secs.~\ref{sec:advantage_refinement} and~\ref{sec:policy_update}. The shared implementation settings are summarized in Table~\ref{tab:implementation_hyperparameters}.

\begin{table}[ht]
    \centering
    \caption{\textbf{Implementation hyperparameters.} }
    \label{tab:implementation_hyperparameters}
    \setlength{\tabcolsep}{3.5pt}
    \renewcommand{\arraystretch}{1.08}
    \begin{tabular}{@{}p{0.70\columnwidth}p{0.20\columnwidth}@{}}
        \toprule
        Hyperparameter & Value \\
        \midrule
        \textit{Policy \& Advantage Training} \\
        Input image resolution & $224 \times 224$ \\
        Action horizon & 20 \\
        Optimizer & AdamW \\
        Policy learning rate & $1.0 \times 10^{-5}$ \\
        Advantage learning rate & $5.0 \times 10^{-5}$ \\
        Learning rate schedule & Cosine\\
        Warm-up steps & 500 \\
        Batch size & 128\\
        Weight decay & $1.0 \times 10^{-10}$ \\
        Update steps per round & 5000\\
        \midrule
        \multicolumn{2}{@{}l}{\textit{Online collection and ACBC}} \\
        Number of stochastic policy samples $N$ & 10\\
        Energy score weight $\lambda_p,\lambda_R,\lambda_g$ & 0.5, 0.25, 0.25 \\
        Advantage-evaluation interval $K$ & 50\\
        Data mixture ratio $\alpha$ & 0.5\\
        Positive-advantage fraction $\eta$ & 0.3\\
        \bottomrule
    \end{tabular}
\end{table}

\subsubsection{Results and Analysis}
\begin{figure*}[t]
    \centering
    \includegraphics[width=\textwidth]{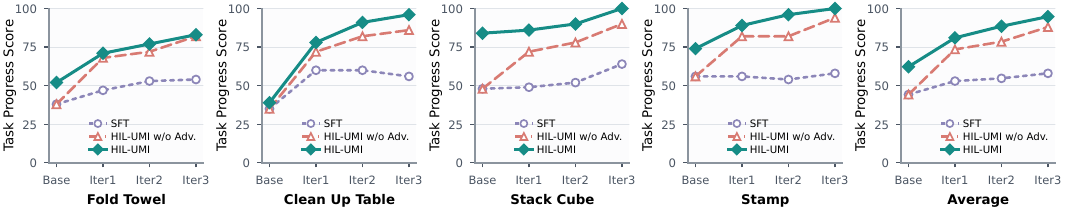}
    \caption{\textbf{Real-world Experiment Results.} We measure the task progress score (TPS) across post-training rounds on four long-horizon or high-precision real-world tasks for SFT, \method without advantage, and \method. We report the average over 4 tasks on the far right.}
    \label{fig:main_results}
\end{figure*}

Figure~\ref{fig:main_results} compares HIL-UMI with SFT under the same per-round data budget. Across all four tasks, SFT yields only limited improvement, whereas HIL-UMI improves consistently throughout post-training. This suggests that simply collecting additional nominal demonstrations is insufficient to reliably address the states encountered by the current policy.

In contrast, HIL-UMI explicitly targets policy-specific OOD regions during data collection and further exploits the collected data through advantage-conditioned policy updates. As a result, the same collection budget is concentrated on supervision that is more relevant to the policy's current weaknesses. The consistent gains on both long-horizon and precise tasks indicate that this strategy provides a more effective use of additional human demonstrations than SFT.

\subsection{Ablation Experiments}
\subsubsection{Ablation on Advantage}
We remove the advantage model in \method, and replace the ACBC update with simple finetuning on the mixed dataset. As shown in Figure ~\ref{fig:main_results}, \method without advantage shows significant performance drop, because it fails to label policy data by utility and treat the task-progressing and suboptimal examples as the same.

\subsubsection{Ablation on Data Collection Thresholds}
We study the sensitivity of the two online collection triggers on Stack Cube, by varying one threshold at a time while fixing the other at the selected setting, $(\tau_P,\tau_A)=(1.2,0.2)$.  

\begin{table}[ht]
    \centering
    \caption{\textbf{Ablation of online collection thresholds.}}
    \label{tab:collection_threshold_ablation}
    \setlength{\tabcolsep}{4.2pt}
    \renewcommand{\arraystretch}{1.08}
    \begin{tabular}{ccccc}
        \toprule
        $(\tau_P,\tau_A)$ & Base & Round 1 & Round 2 & Round 3 \\
        \midrule
        $(1.2,0.2)$ & 84 & \textbf{86} & \textbf{90} & \textbf{100} \\
        $(2.0,0.2)$ & 84 & 76          & 86          & 96          \\
        $(0.5,0.2)$ & 84 & 80          & 72          & 90          \\
        $(1.2,0.3)$ & 84 & 76          & 82          & 96          \\
        $(1.2,0.1)$ & 84 & 82          & 82          & 92          \\
        \bottomrule
    \end{tabular}
\end{table}

As shown in Table~\ref{tab:collection_threshold_ablation}, moderate thresholds consistently perform best.
For the policy OOD detector, an overly permissive threshold collects less informative states where the policy already agrees reasonably well with the human, whereas an overly conservative threshold can miss useful policy failures.
The advantage detector exhibits a similar trade-off: excessive triggering introduces redundant refinement data, while insufficient triggering misses informative estimator errors.
Overall, both detectors benefit from balancing \emph{coverage} and \emph{selectivity}.
HIL-UMI also consistently outperforms SFT across the tested settings, indicating that its improvement is not sensitive to a narrowly tuned threshold.

\subsection{Collection Time Efficiency Experiments}
\begin{figure}[!t]
    \centering
    \includegraphics[width=\columnwidth]{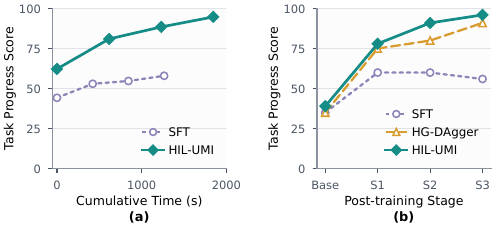}
    \caption{\textbf{Collection-time efficiency and comparison with HG-DAgger.}
    (a) We report the average TPS and corresponding  time cost over 4 real-world tasks for SFT and HIL-UMI collection. (b) We compare SFT, HG-DAgger and HIL-UMI on Clean Up Table and report the TPS for each round.}
    \label{fig:collection_time_efficiency}
\end{figure}

We compare the collection-time efficiency of SFT and \method. Figure~\ref{fig:collection_time_efficiency}(a) plots
the four-task mean TPS against the cumulative mean collection time across
post-training rounds.

Although the targeted collection for HIL-UMI takes longer per recorded frame (Table~\ref{tab:collection_time_per_frame}), its mean TPS rises steadily while SFT plateaus and temporarily regresses. These experiment results indicate that the additional online selection overhead is therefore offset by collecting around policy-specific blind spots and prioritizing task-progressing supervision.
\begin{table}[ht]
    \centering
    \caption{\textbf{Collection Efficiency Comparison.}}
    \label{tab:collection_time_per_frame}
    \setlength{\tabcolsep}{3.6pt}
    \renewcommand{\arraystretch}{1.08}
    \begin{tabular}{lcc}
        \toprule
        Task & SFT (ms/frame) & \method (ms/frame) \\
        \midrule
        Fold Towel      & 69.43 & 91.92  \\
        Clean Up Table  & 41.70 & 73.40  \\
        Stack Cube      & 77.04 & 89.89  \\
        Stamp           & 64.30 & 101.02 \\
        \bottomrule
    \end{tabular}
\end{table}
\subsection{Comparison with HG-DAgger}

Figure~\ref{fig:collection_time_efficiency}(b) compares real-robot HG-DAgger with \method on Clean Up Table under the same per-stage budget. \method consistently achieves higher TPS across all stages and finishes with a TPS approximately five points higher than that of HG-DAgger. This improvement is consistent with the design of ACBC, which allows the policy to favor high-advantage behaviors at inference time while still leveraging suboptimal data during training. In addition, \method is substantially more efficient in data collection: HG-DAgger requires 412.99~ms per frame, which is 5.63$\times$ the 73.40~ms per frame required by \method. This gap demonstrates the collection efficiency gain from avoiding robot rollouts.

\section{Conclusion}
We introduced HIL-UMI, a human-in-the-loop framework for iterative VLA post-training without robot rollouts. During handheld UMI demonstrations, HIL-UMI targets policy OOD states and iteratively refines an advantage estimator to label collected data for advantage-conditioned behavior cloning. Across four long-horizon and precise manipulation tasks, HIL-UMI consistently outperformed SFT, and it also achieved significantly higher collection efficiency than HG-DAgger. 

Future work will develop \method into a distributed post-training system in which operators collect policy-guided UMI data concurrently across locations. Therefore, \method points toward scalable VLA post-training driven by distributed human data without repeated robot deployment.

\section{Acknowledgment}
We thank Zhewei Gui and Junhan Wang for their insightful discussion. This research was supported by Beijing Natural Science Foundation (26L080330) and National Natural Science Foundation of China (62376006).


\bibliography{ref}

\end{document}